# Multilingual Cross-domain Perspectives on Online Hate Speech

Tom De Smedt, Sylvia Jaki, Eduan Kotzé, Leïla Saoud,
Maja Gwóźdź, Guy De Pauw and Walter Daelemans

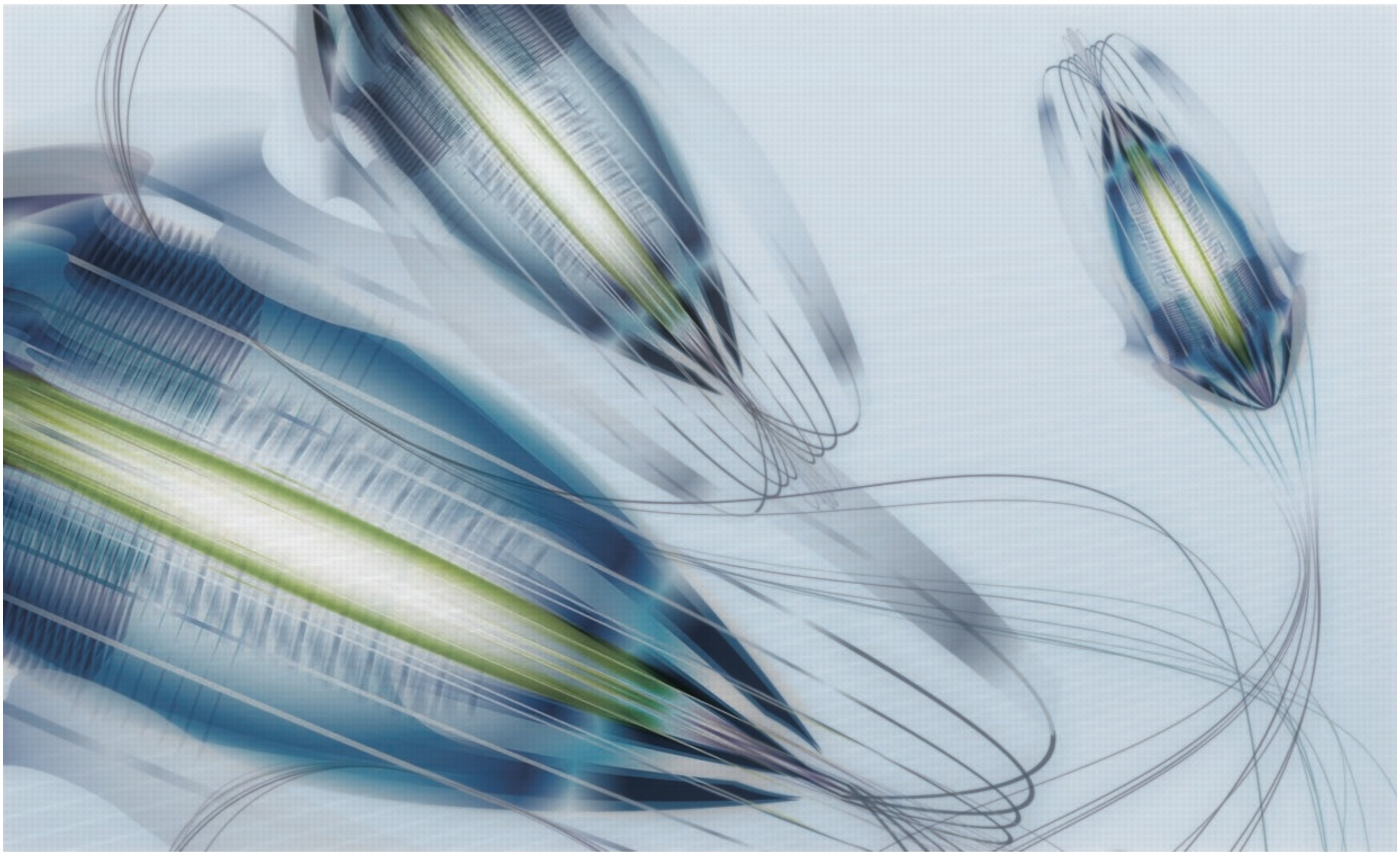

CLiPS
Computational Linguistics & Psycholinguistics
University of Antwerp

# Multilingual Cross-domain Perspectives on Online Hate Speech

Computational Linguistics & Psycholinguistics Research Center

## Authors

**Dr. Tom De Smedt**
CLiPS Research Center, University of Antwerp (BE)
tom.desmedt@uantwerpen.be

**Dr. Sylvia Jaki**
Dept. of Translation & Specialized Communication, University of Hildesheim (DE)
jakisy@uni-hildesheim.de

**Dr. Eduan Kotzé**
Dept. of Computer Science and Informatics, University of the Free State (ZA)
KotzeJE@ufs.ac.za

**Leïla Saoud**
Dept. of Computer Science, University of Leuven (BE)
ls@leilasaoud.com

**Maja Gwóźdź**
Dept. of Mathematics, University of Munich (DE)
maja.gwozdz.mkg33@gmail.com

**Dr. Guy De Pauw**
Textgain
guy@textgain.com

**Prof. Dr. Walter Daelemans**
CLiPS Research Center, University of Antwerp (BE)
walter.daelemans@uantwerpen.be

# Multilingual Cross-domain Perspectives on Online Hate Speech

Tom De Smedt,[1] Sylvia Jaki,[2] Eduan Kotzé,[3] Leïla Saoud,[4]
Maja Gwóźdź,[5] Guy De Pauw,[6] Walter Daelemans[1]

[1] University of Antwerp, [2] University of Hildesheim, [3] University of the Free State,
[4] University of Leuven, [5] University of Munich, [6] Textgain

**Abstract.** In this report, we present a study of eight corpora of online hate speech, by demonstrating the NLP techniques that we used to collect and analyze the jihadist, extremist, racist, and sexist content. Analysis of the multilingual corpora shows that the different contexts share certain characteristics in their hateful rhetoric. To expose the main features, we have focused on text classification, text profiling, keyword and collocation extraction, along with manual annotation and qualitative study.

## 1 Introduction

Hate crimes have been on the rise,[1 2 3] and especially online social media are believed to act as a propellant for **polarization** and **radicalization**. Previous studies have argued that online social media can function as "echo chambers" (Colleoni, Rozza & Arvidsson, 2014), lending themselves to the expression of more radical views than face-to-face interaction. As such, hate speech is perceived to infiltrate various types of (mainly) political discourse online.

Hate speech is defined in the Encyclopedia of the American Constitution as "communication that disparages a person or a group on the basis of some characteristic such as race, color, ethnicity, gender, sexual orientation, nationality, religion, or other characteristic" (Nockleby, 2000). It is illegal according to Article 20 of the International Covenant on Civil and Political Rights.[4] However, in the US, freedom of speech is protected by the First Amendment of the United States Constitution and there is no exception that prohibits hate speech, unless it constitutes incitement to imminent crime (*Brandenburg v. Ohio*, 395 U.S. 444, 1969). In the EU, the Code of Conduct on countering illegal hate speech online[5] defines hate speech as "the public incitement to violence or hatred directed to groups or individuals on the basis of certain characteristics, including race, colour, religion, descent and national or ethnic origin".

---

[1] http://www.reuters.com/article/u-s-hate-crimes-rise-for-second-straight-year-fbi-idUSKBN1DD2BA
[2] http://www.reuters.com/article/uk-hate-crimes-surge-on-brexit-and-militant-attacks-idUSKBN1CM15E
[3] http://www.reuters.com/article/us-europe-rights-idUSKBN1672AB
[4] http://www.ohchr.org/EN/ProfessionalInterest/Pages/CCPR.aspx
[5] http://www.europa.eu/rapid/press-release_IP-16-1937_en.htm

Hate speech laws are now being considered in several countries such as the UK and France,[6] Germany (Netzwerkdurchsetzungsgesetz)[7] and South Africa (B9–2018).[8] Due to the pervasive nature of hate speech, such laws are often notoriously vague,[9] [10] which can lead to potential problems with the infringement on freedom of speech of private citizens. The response by IT companies has also been rather cautious, for example with Facebook CEO Mark Zuckerberg stating before US Congress that the problem can be solved in "5-10 years".

Hate speech online presents a new challenge for Natural Language Processing (NLP). The aim of this study is to identify common features of hate speech across domains, and to advance automatic detection. We collected 8 social media text corpora for a multilingual and cross-domain study of hate speech, composed of jihadism, right- and left-wing extremism, racism, and sexism (section 2). We provide an overview of the NLP techniques that we have used to gain insight from these resources, in particular using text classification, keyword extraction, collocation extraction (section 3) and stylometry and sentiment analysis (section 4).

# 2 Methods and Materials

## 2.1 Context

Table 1 provides an overview of our target domains, followed by a summary of the social and political context of each domain in respective countries:

| DOMAIN | CONTEXT | LANGUAGE |
|---|---|---|
| JIHADISM | Islamic State (ISIS) propaganda | EN / AR |
| EXTREMISM | far-right activism in Germany | DE |
| EXTREMISM | far-right activism in Belgium & the Netherlands | NL |
| EXTREMISM | political debate in the UK / US / Canada | EN |
| SEXISM | incel subculture (male supremacy) | EN |
| RACISM | far-left activism in South Africa | EN |
| RACISM | far-right activism in France | FR |
| RACISM | racist comments in Belgium | NL |

**Table 1.** Overview of target domains.

[6] http://www.gov.uk/government/news/uk-and-france-announce-joint-campaign-to-tackle-online-radicalisation
[7] http://www.buzer.de/NetzDG_Netzwerkdurchsetzungsgesetz.htm
[8] http://www.justice.gov.za/legislation/hcbill/hatecrimes.html
[9] http://www.hrw.org/news/2018/02/14/germany-flawed-social-media-law
[10] http://www.hrw.org/news/2017/02/21/south-african-move-hate-speech-step-too-far

JIHADISM

During the Syrian Civil War and the rise of the Islamic State (IS, ISIS, ISIL, Daesh) in Iraq and Syria, Salafi jihadists were able to set up an effective propaganda machine on online social media such as Twitter and YouTube (Tomé, 2015), spreading fundamentalist views on Islam, war reporting, execution videos, and terrorist manuals, to incite fear and hatred and to recruit new members (Klausen et al., 2012). For example, on August 19, 2014, a video was released on YouTube showing the beheading of abducted US journalist James Foley, to be met with worldwide condemnation. The EU subsequently announced a new Code of Conduct, and Twitter reported suspending over 125,000 user profiles related to ISIS.[11] Insurgents then moved to anonymous messaging apps such as Telegram (Weimann, 2016).

EXTREMISM (EU)

In the wake of the Syrian crisis, high numbers of refugees arrived in the EU. For example, the number of refugees arriving in Germany more than doubled in 2015 (Eurostat, 2017). During this time, the country also witnessed a number of violent incidents with refugees, such as the 2015 New Year's Eve sexual assaults and the December 2016 Berlin truck attack. These events sharply polarized the public opinion (YouGov, 2016), which became clear during the 2017 German federal election, with a 12.6% victory for the far-right AfD. Since then, there has been a surge of German far-right propaganda on social media (Davey & Ebner, 2017), correlating with increased violence towards refugees (Müller & Schwarz, 2017). For example, after a stabbing involving an Iraqi and a Syrian in Chemnitz on August 27, 2017, hundreds of far-right protesters took to the streets, "fueled by fake news", "hunting down" immigrants.[12]

In Belgium, authorities have struggled with the prominent role of Belgian foreign fighters in Syria and Iraq, and their involvement in the November 2015 Paris attacks and the 2016 Brussels bombings (Van Ostaeyen, 2016), incidentally with US presidential candidate Donald Trump calling Brussels a "hellhole".[13] In 2010, the Belgian federal election already revealed a deep divide between the left wing and the right wing, which led to a year-long government formation crisis. The immigration debate, in particular concerning Muslims, is perceived to further polarize opinions (see Torrekens, 2015: 161). To illustrate this, on August 20, 2018, Belgian security services issued a remarkable press statement warning that far-right militias patrolled the streets "to protect citizens" and that "violent incidents could not be ruled out".[14]

---

[11] http://blog.twitter.com/official/en_us/a/2016/combating-violent-extremism.html
[12] http://www.dw.com/en/german-state-official-fake-news-fueled-chemnitz-riots/a-45263589
[13] http://www.politico.eu/article/donald-trump-brussels-is-like-a-hellhole
[14] http://www.sudinfo.be/id70744/article/2018-08-20/selon-la-surete-de-letat-lextreme-droite-se-renforce-en-belgique-plusieurs

The EU Terrorism Situation and Trend Report (Europol, 2018) notes that violent right-wing extremism is expanding, “partly fuelled by fears of a perceived Islamization of society and anxiety over migration”, particularly in the case of the far-right Identitarian movement, which currently has branches in Germany, Austria, France, the Netherlands, and the US.

EXTREMISM (US)

In the US, a nation deeply divided by liberal and conservative views (Westfall et al., 2015), new President Donald Trump has successfully tapped into the concerns of conservative “angry white men” (Kimmel, 2017), for example by linking immigration to crime, globalization to unemployment, and politics to elitism.[15] His “post-truth” Twitter messages have been called sexist, racist, and contagious (Ott, 2017: 64), and have contributed to a strained relationship with the press, each accusing the other of proliferating fake news. A recent study has shown that, on Facebook, fake news stories tend to favor Trump (Allcott & Gentzkow, 2017).

Trump has repeatedly framed the press as an “Enemy of the People” (Figure 1), drawing criticism but also inflaming public sentiment. To illustrate this, in August 2018, the New York Times received a threatening voicemail declaring: “You’re the problem. You are the enemy of the people. And although the pen might be mightier than the sword, the pen is not mightier than the AK-47”. The Boston Globe received another call,[16] saying: “You’re the enemy of the people, and we’re going to kill every fucking one of you. […] Still there faggot? […] I’m going to shoot you in the fucking head later today, at 4 o’clock.”

The man charged with threatening the Boston Globe stated that he will continue to do so until the newspaper stops its “treasonous” attacks on Trump.

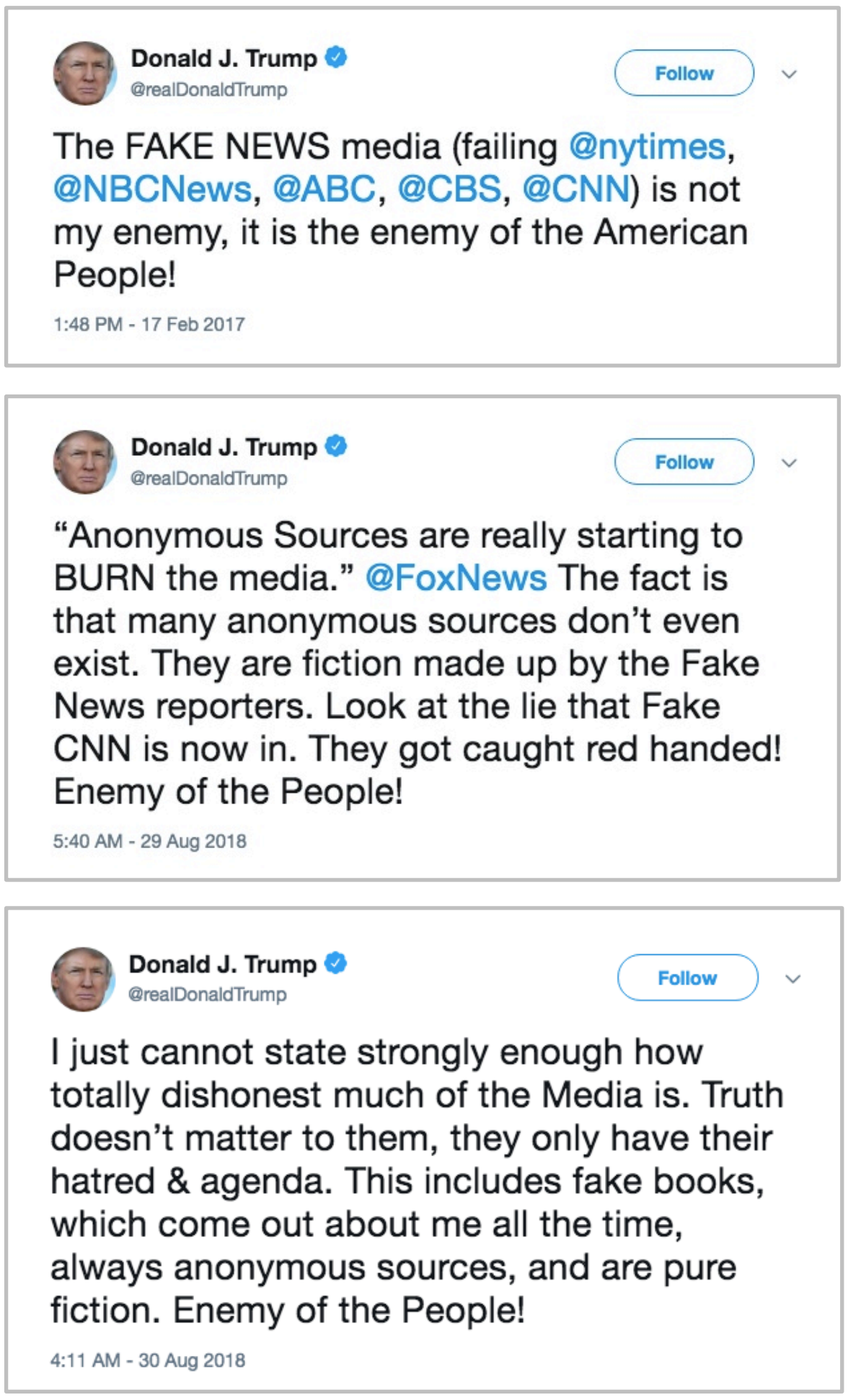

**Figure 1.** Example tweets by Donald Trump.

[15] http://time.com/4566304/donald-trump-revenge-of-the-white-man
[16] http://time.com/5383632/donald-trump-editorial-death-threat

SEXISM

Two recent killing sprees in the US and Canada, the 2014 Isla Vista killings and the 2018 Toronto van attack, have brought the online male supremacist movement to the attention of the media. In particular, both perpetrators self-identified as *incels* or involuntary celibates, a subculture of “angry white men” that blame women for depriving them of sexual contacts, using violent, misogynist, homophobic, and racist vernacular (Ging, 2017). For example, one user on the Incels.me forum states: “Even if it’s shooting up a school, don’t sit back and do nothing”. This shows how hate speech is not necessarily always politically motivated, even though ties between male supremacy and the far-right have been observed (Nagle, 2017).

RACISM (EU)

In France, anti-immigration and far-right political parties have steadily gained in popularity (Golder, 2016). In 2017, the Rassemblement National (formerly Front National) qualified for the second round of the presidential election and received 33.9% of the votes. The party has advocated against immigration, linking it to terrorism and crime (Davies, 2012) and fueling xenophobic sentiment in the public opinion, which has been perceived to contribute to the rise of racism on social media (Benveniste & Pingaud, 2016).

In Belgium, the far-right Vlaams Belang rose steadily to 19% of the votes in the 2007 federal election, but dropped to 5.8% in 2014, with voters turning to the right-wing N-VA (32.2%), in part because the N-VA is not subject to *cordon sanitaire* (an agreement between all parties not to form a coalition with Vlaams Belang). On August 30, 2018, a message was plastered on a mosque in Leuven stating: “Vote N-VA, brown people out”, along with Nazi symbolism.[17]

RACISM (AF)

In South Africa, the left-wing ANC has 62.15% of the votes and the far-left EFF 6.35% (third-largest party). Recently, the government has began advocating for more stringent land reform regulations (i.e., expropriation without compensation), to redistribute land owned by white Afrikaners (~72%) to black farmers. Commentators have subsequently noted an increase of online hate speech targeting white people (e.g., Steward, 2016; Khoza, 2017). At the same time, the country has experienced a number of farm attacks that may have been racially motivated (see Kerkvliet, 2017), while new hate speech regulations (Bill B9-2018) have been criticized for violating freedom of expression.[18]

[17] http://www.vrt.be/vrtnws/en/2018/08/31/racist-slur-on-leuven-mosque-alle-bruine-buite

[18] http://www.hrw.org/news/2017/02/21/south-african-move-hate-speech-step-too-far

## 2.2 Corpora

Table 2 provides an overview of the collected corpora:

| DOMAIN | REGION | LANGUAGE | PERIOD | GENRE | SAMPLE |
|---|---|---|---|---|---|
| JIHADISM | Iraq & Syria | EN / AR | 2014-2016 | Twitter | 50,000 |
| EXTREMISM | Germany | DE | 2017-2018 | Twitter | 55,000 |
| EXTREMISM | Belgium & the Netherlands | NL | 2017-2018 | Twitter | 30,000 |
| EXTREMISM | UK / US / Canada | EN | 2017-2018 | Twitter | 7,500 |
| RACISM | South Africa | EN | 2017-2018 | Twitter | 10,000 |
| RACISM | France | FR | 2017-2018 | Facebook | 10,000 |
| RACISM | Belgium | NL | 2015-2016 | Facebook | 5,000+ |
| SEXISM | - | EN | 2017-2018 | Incels.me | 65,000 |

**Table 2.** Overview of collected corpora.

Combined, the corpora cover jihadism (~20%), extremism (40%), racism (10%) and sexism (30%), although not every text message in every corpus necessarily constitutes hate speech. Most text messages are written in English (~45%), German (25%) and Dutch (15%), which are linguistically similar, complemented with messages written in French (5%), Arabic (5%) and assorted other languages (5%). Following is a summary of our data collections methods.

JIHADISM (EN/AR)

In the period from October 2014 to December 2016, we collected about 50,000 tweets posted by 350+ jihadists (De Smedt, De Pauw & Van Ostaeyen, 2018), in the wake of 10 terrorist attacks (e.g., Berlin, Brussels, and Paris attacks). For example: "May the streets of France be filled with the blood of these filthy kuffar [...]" (January 2015). The profiles were identified manually using a combination of search keywords (e.g., *kuffar*, unbelievers), cues in the profile usernames (e.g., *muhajir*, foreign fighter) and profile pictures (e.g., masks, weapons, or images of lions, which symbolize bravery). The tweets posted by these profiles were then automatically collected using the Pattern toolkit for the Python programming language (De Smedt & Daelemans, 2012a). About 40% of the tweets are in English, 30% are in Arabic and 5% are in French. For comparison, we also collected a companion corpus of 50,000 tweets by news agencies, imams, muslimas, etc., that talk about the Syrian crisis without overt bias.

EXTREMISM (DE)

In the period from August 2017 to April 2018 we collected about 55,000 tweets posted by 100+ German right-wing extremists (Jaki & De Smedt, 2018). For example: “Das Pack gehört in Straflager” (the rabble belongs in the camps, March 2018). The profiles were identified manually using a combination of search keywords (e.g., *Neger*, nigger), cues in the profile usernames (e.g., *59* = *EI*, shorthand for *Eil Itler*) and profile pictures (e.g., Hitler moustache parodies, knight crusaders, blue-eyed wolves). The tweets posted by these profiles were then automatically collected using Pattern. For comparison, we also collected a companion corpus of 20,000 tweets by well-known German politicians, who we expect not to post hate speech, and 30,000 random German tweets. About 35% of the data is publicly available in the POLLY corpus (De Smedt & Jaki, 2018a), along with other (anonymized) political German tweets.[19]

EXTREMISM (NL)

In the period from September 2017 to August 2018, we used a manual set of high-precision Dutch keywords (e.g., *makakken*, offensive word for Moroccan immigrants) to continuously collect tweets and track trends in the data. This resulted in a corpus of about 30,000 tweets, and trending insights such as the spread of new offensive words such as *kansparel* (roughly: pearl of opportunity). For example: “Dank u Merkel, alweer een kansparel van u” (thanks Merkel, another one of your immigrants, August 2018). Figure 2 shows a timeline of tweets collected per month. In general, Dutch hate speech is on the rise. We can also correlate spikes in the data to real-world events. For example, on March 16, 2017, we collected 3x as many hate tweets as on other days in March, after it became known that a convicted Belgian foreign fighter could not be extradited because of human rights concerns.[20]

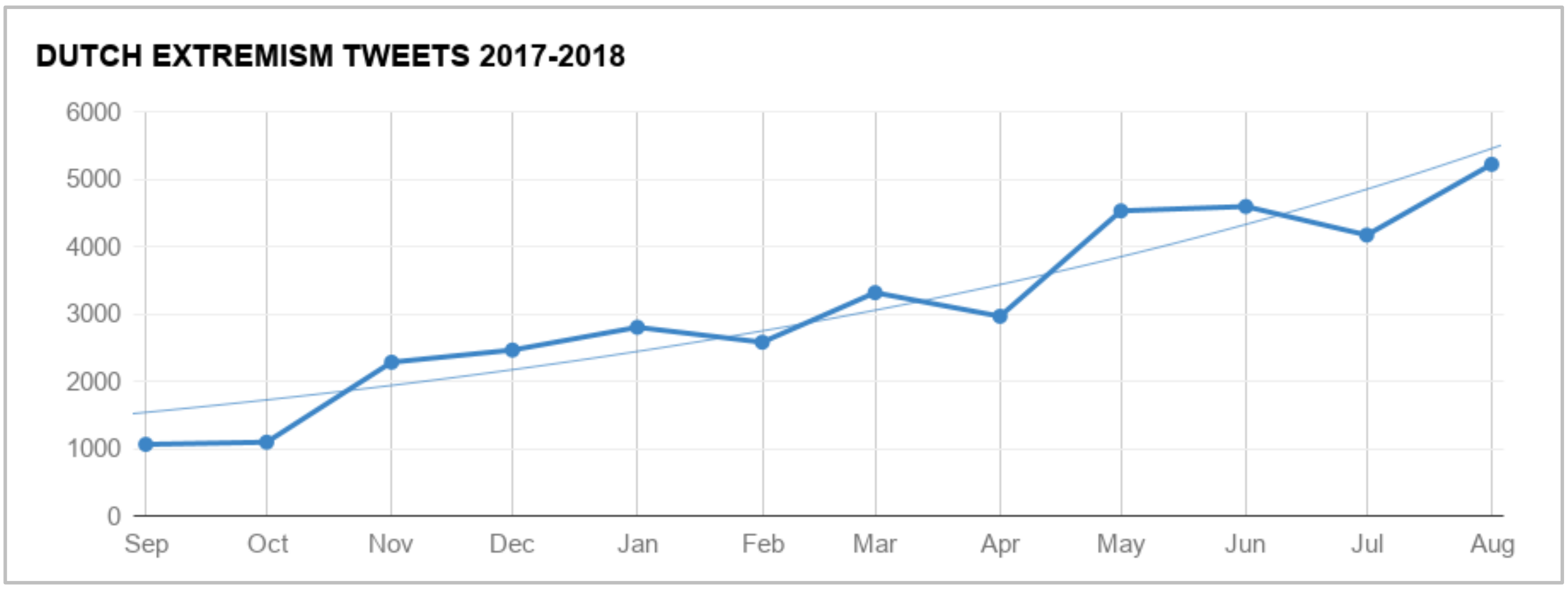

**Figure 2.** Timeline of collected Dutch extremism tweets.

[19] https://docs.google.com/spreadsheets/d/1c5peNMjt24U0FcEMSj8gD_JjzumqXTWbPWa_yb2nNt0

[20] https://www.tijd.be/politiek-economie/belgie/federaal/francken-mag-veroordeelde-syriestrijder-niet-uitwijzen/9992783.html

EXTREMISM (EN)

During Google Summer of Code 2018, we collected about 7,500 tweets (Gwóźdź, 2018) with political debate in the US (35%), the UK (35%), and Canada (20%), where each tweet has been manually annotated with keywords, speech acts (~10% directives), profanity (~15%), polarity (45% negative vs 10% positive), mood (15% angry vs 5% happy), gender (50/50 men vs women), political bias (50/50 left-wing vs right-wing), and metaphors (<1%; see also Sulis et al., 2016). For example, "@realDonaldTrump: You are a pathetic, disgusting excuse for a human being" is labeled as left-wing, negative, and agitated, with *pathetic* and *disgusting* marked as profanity. The data is publicly available in Google Sheets as the MAGA corpus.[21]

RACISM (FR/NL)

In February 2018, we collected about 60,000 public posts from Facebook newspaper pages such as La Dernière Heure and Le Figaro, using the Graph API (Saoud, 2018). For example: "Ton Coran tu te le mets ou je pense, vous n'etes que des barbares" (put your Quran where I'm thinking, you are nothing but barbarians). Over 10,000 of these were labeled by multiple annotators for racism (high agreement, $K$=0.92), which constitutes about 15% of the subset. These were then used to extract a set of high-precision French racial slurs (e.g., *crouille*) for continuous monitoring. We also annotated over 5,000 posts in Dutch for racism ($K$=0.60), collected from Facebook pages that are likely to attract racist content (Tulkens et al., 2016).

RACISM (EN/AF)

In the period from September 2017 to March 2018, we collected over 10,000 tweets with the hashtag #Orania (Kotzé & Senekal, 2018). For example: "That #Orania shit place must be burnt down, we want them back in Europe" (January, 2018). Orania is a minority community in South Africa that seeks to preserve Afrikaner culture and language amongst a majority of black African population. The community has been the subject of polarized political debate, often being accused by the far-left of upholding racial segregation.

SEXISM (EN)

In April 2018, we collected about 65,000 messages by 1,250 users in 3,500 discussion threads on the Incels.me forum (Jaki et al., 2018), after a 25-year-old incel killed 10 in the Toronto van attack. The corpus contains instances of misogyny (30%), homophobia (15%) and racism (3%). For example: "The slut DESERVED the murder 100%" (April, 2018). For comparison, we also collected a companion corpus of 40,000 paragraphs of Wikipedia texts (moderated for neutrality) and 10,000 random English tweets (to account for internet slang).

[21] https://docs.google.com/spreadsheets/d/1mFV7uIEbMQ9LyaLRLQc-c0zVfKFn0CY_DakHSYWyPNg

## 2.3 Methodology

We applied a combination of known “superficial” NLP techniques (e.g., classification models based on character trigrams, keyword extraction by word frequency, sentiment analysis using dictionaries) on the collected corpora. The study was explorative, meaning that we did not apply all techniques to all datasets. Most of the work has been trial-and-error.

# 3 Results

## 3.1 Text Classification

Machine Learning (ML) is a subfield of AI that uses statistical techniques to train systems that “learn by example”. For example, given 10,000 HATE messages and 10,000 SAFE messages, the system will automatically discover that words or word combinations such as *filthy pigs* occur more often in hateful messages, which can then be used as cues to detect whether other (unknown) messages are hateful or not. In the process, each training example is mapped to a vector of features with weights. The features could be the words in a message for example, and the weights could be word count. In our case, we used character trigrams as features, where *filthy pigs* is mapped to { *fil*, *ilt*, *lth*, *thy*, ... }, so that the system can deal with spelling errors, word endings, etc., more efficiently (e.g, *filth* and *filthy* will have several overlaps). We usually also included character unigrams and bigrams, and word unigrams and bigrams, which often raises the predictive performance by a few percent.

Table 3 provides an overview of the predictive performance for each domain, measured by recall (cf., quantity, how many hateful messages are discovered) and precision (cf., quality, how many messages predicted as hateful are *really* hateful), using 10-fold cross validation:

| DOMAIN | CLASSIFICATION MODEL | TASK | LANGUAGE | PRECISION | RECALL |
|---|---|---|---|---|---|
| JIHADISM | support vector machine | HATE vs SAFE | EN / AR | 82% | 82% |
| EXTREMISM | single-layer perceptron | HATE vs SAFE | DE | 84% | 84% |
| EXTREMISM | decision trees | LEFT vs RIGHT | EN | 82% | 81% |
| RACISM | support vector machine | HATE vs SAFE | FR | 84% | 83% |
| RACISM | support vector machine | HATE vs SAFE | NL | 39% | 70% |
| SEXISM | single-layer perceptron | HATE vs SAFE | EN | 93% | 92% |

**Table 3.** Overview of predictive performance (precision & recall).

For example, the LIBSVM algorithm (Chang & Lin, 2011) yields 82% $F_1$ score (the mean of precision and recall) for jihadism detection, while Perceptron (Collins, 2002) yields 84% $F_1$ for German right-wing extremism and 92.5% for sexism. We achieved the best results with a deep learning Convolutional Neural Network (CNN; Kim, 2014) for sexism: about 95%. This shows that it is useful to discuss automatic detection systems for hate speech. However, such systems are also problematic once deployed "in the wild", because of their high error rate. For example, 80% accuracy means that 2 out of 10 SAFE messages are labeled HATE, or vice versa, raising ethical concerns. Two other main issues are scalability and interpretability:

**Scalability**. In a technical study on automatic detection of offensive language (De Smedt & Jaki, 2018b), we discuss how in-domain performance can be misleading when systems are applied to out-of-domain data. Classification models may be prone to overfitting, memorizing training examples instead of learning general trends, and consequently perform poorly during cross-domain evaluation. But cross-domain scalability is essential, since perpetrators will use countermeasures against identification (Berger & Perez, 2016), for example by adapting their language use. With the rising interest in addressing hate speech online, some systems have reported up to 99% accuracy.[22] This may be somewhat optimistic, since even policy makers and legal experts do not agree on what exactly constitutes hate speech across domains.

**Interpretability**. The best results are typically achieved using the most recent deep learning techniques. However, the decision-making process of these systems is often also more difficult to explain ("black box"), raising legal concerns. In this regard, systems built for deployment should adopt a multipronged approach, where the best classification models are accompanied by explanatory tools, such as decision trees to inspect the decision-making process, keyword dictionaries to highlight words for human moderators, and visualizations of language use like word clouds and word trees. Figure 3 shows a decision tree trained on the MAGA corpus:

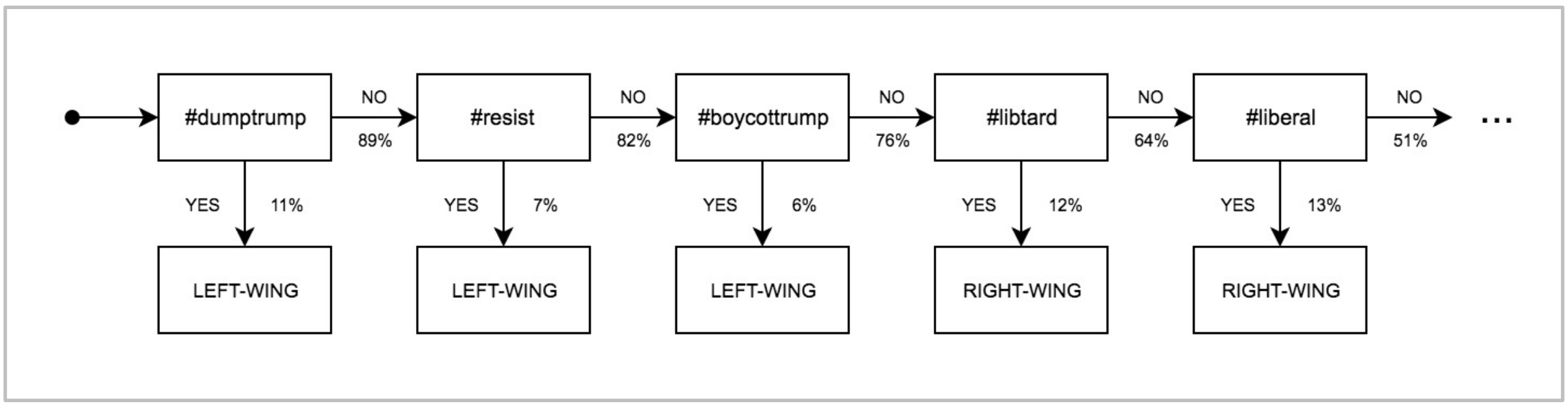

**Figure 3.** Decision tree trained on the MAGA corpus (visualization).

[22] https://i-hls.com/archives/81392

Figure 4 shows the source code to train the decision tree on left-wing vs right-wing tweets:

```
# coding: utf-8
# http://github.com/textgain/grasp
from grasp import download, tmp, csv, DecisionTree, wc, tokenize, kfoldcv
# Google Sheets can be downloaded as a CSV-formatted string:
MAGA  = 'https://docs.google.com/spreadsheets/d/%s/gviz/tq?tqx=out:csv&sheet=%s'
MAGA %= '1mFV7uIEbMQ9LyaLRLQc-c0zVfKFn0CY_DakHSYWyPNg', 'FULL'
# Each training example will be mapped to a {word: count}-dict:
def v(s):
    return wc(tokenize(s)) # "SAD!" => {'sad': 1, '!': 1}
data = []
with tmp(download(MAGA, cached=True)) as f:
    for r in csv(f.name)[1:]:
        if r[8].startswith(('left', 'right')):
            data.append((v(r[2]), r[8])) # (vector, label)
# Calculate precision & recall:
P, R = kfoldcv(DecisionTree, data, k=3, min=100)
print(P)
print(R)
```

**Figure 4.** Decision tree trained on the MAGA corpus (source code).

The source code uses the Grasp toolkit for the Python programming language.[23] It downloads the MAGA corpus from Google Sheets, maps each tweet to a word → count vector, each with a LEFT-WING or a RIGHT-WING label, and trains a decision tree with a minimum leaf size of 100. It then computes precision and recall, using 3-fold cross-validation. We get about 80% $F_1$ score. As shown in Figure 3, nearly 50% of the labels was learned by hashtags. For example, any tweet that contains *#libtard*, but not *#dumptrump*, *#resist* or *#boycottrump*, will now be predicted as RIGHT-WING. The decision-making sounds plausible, however this model is not very scalable and may be outdated soon, since popular hashtags change all the time.

## 3.2 Keyword Extraction

Another way to interpret large volumes of text that worked well for us is keyword extraction, using feature selection techniques (Liu & Motoda, 2007). Given a HATE and SAFE corpus, we can use word count with a chi-square test to expose significantly biased words, and posterior probability to find out in what set a word occurs more often. For example, in the JIHADISM corpus, the word *kuffar* (unbeliever) is significantly biased ($p<0.05$) and occurs in about 3% of HATE tweets (1,500x) and 0.05% of SAFE tweets (25x). It is a useful cue for hate speech.

[23] http://github.com/textgain/grasp

By examining keywords across the collected corpora, we can identify a number of general mechanisms prevalent in all domains:

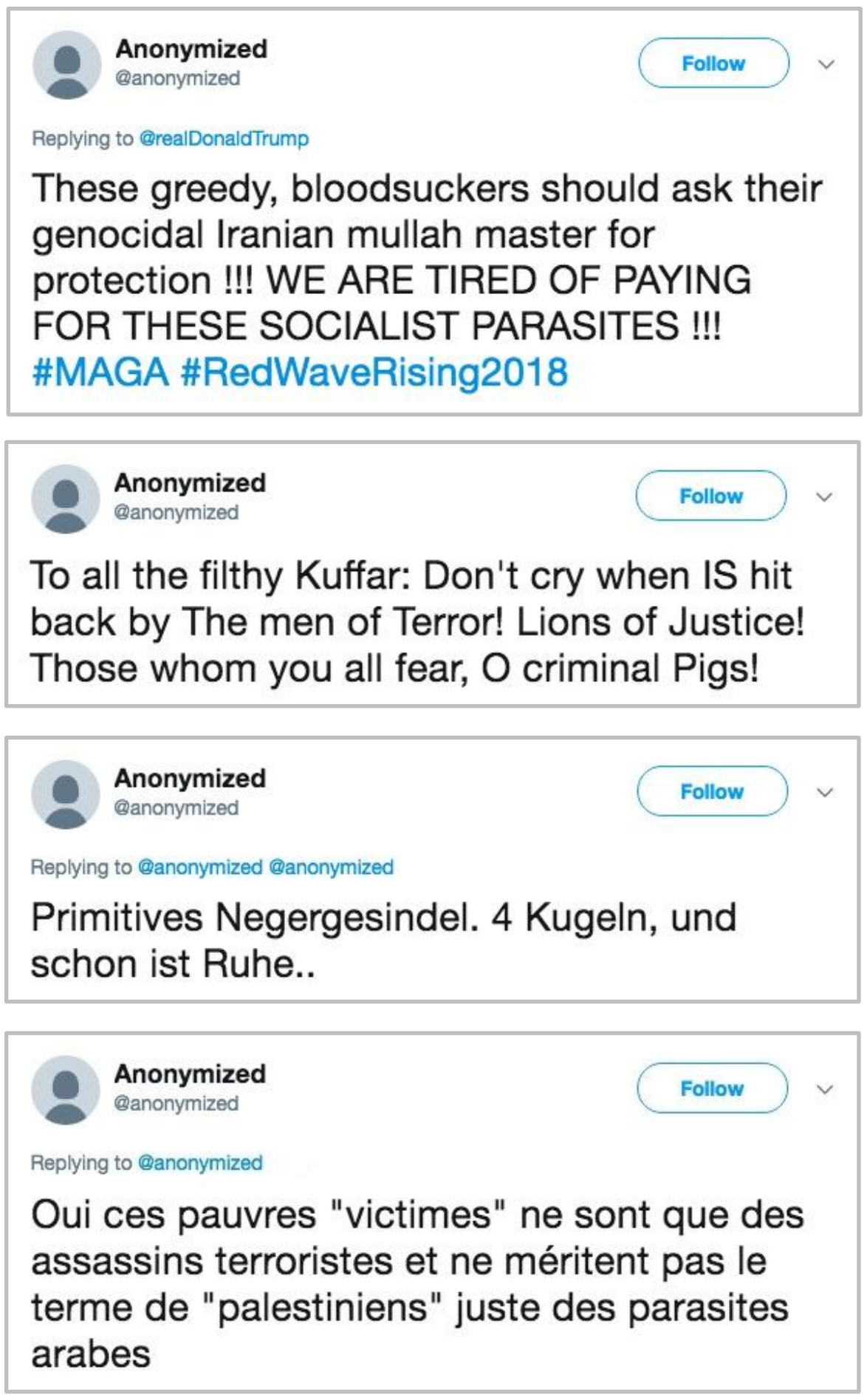

**Figure 5.** Examples of dehumanization.

**Slang**. In-group expressions that are used to promote group identity by categorizing adversaries, such as *coconut* (Muslim apostate; brown outside, white inside), *gutmensch* (do-gooder), *stacy* (attractive woman). Outsiders may not be aware of their meaning, but they will be used frequently and hence also stand out with keyword extraction.

**Slurs**. Disparaging expressions that are used to target adversaries by race, gender, sexual orientation, religion (e.g., *nigger*, *little bitch*, *infidel*, *faggot*), or to defame political opponents (e.g., *drunkard*, *dumbass*). Typically, a small variety is used repeatedly as a means of propaganda, defamation, or incitement.

**Dehumanization**. Disparaging expressions that are used to debase adversaries, to legitimize their removal or their extermination. Typically, this involves comparing individuals or groups to animals that are seen as unclean (e.g., *dog*, *pig*, *vermin*), describing them as waste (e.g., *scum*, *shit*, *trash*) or, predominantly in the case of misogyny, as objects (e.g., *cum dumpster*). Figure 5 shows a number of examples of dehumanization.

**Stereotyping**. Generalizing expressions that are used to frame heterogeneous individuals as a homogenous group, to argue that they are all accountable for reprehensible behavior based on a specific incident (e.g., a crime report). This includes using collective nouns (e.g., *horde*, *rabble*, *wave*) to describe immigrants, using social pejoratives (e.g., *bum*, *moron*, *whore*), and referring to historical conflicts (e.g., *barbarian*, *crusader*, *nazi*).

**Incitement** to crime and violence occurs less frequently, but it is also potentially illegal. It usually involves verbs expressing aggression (e.g., *fight*, *kill*, *rape*), directives (*arm yourself*), and indirect speech acts (*when will you learn?*) that are masked calls-to-action.

Table 4 provides an overview of significantly biased HATE words ($p$<0.05, non-exhaustive):

| DOMAIN | SLANG | SLURS | DEHUMANIZING | STEREOTYPING | INCITEMENT |
|---|---|---|---|---|---|
| JIHADISM | *brother*<br>*coconut*<br>*lion* | *kafir*<br>*murtadd*<br>*rawafid* | *dogs*<br>*pigs*<br>*scum* | *crusader*<br>*hypocrite*<br>*prostitute* | *burn*<br>*destroy*<br>*slaughter* |
| EXTREMISM (DE) | *gutmensch*<br>*linksgurke*<br>*merkelgast* | *muselmann*<br>*nafri*<br>*neger* | *gesindel*<br>*pack*<br>*parasiten* | *barbaren*<br>*horden*<br>*salafisten* | *abschieben*<br>*bewaffnen*<br>*kämpfen* |
| EXTREMISM (EN) | *brexiteer*<br>*libtard*<br>*snowflake* | *bitch*<br>*fucker*<br>*soyboy* | *maggot*<br>*puppet*<br>*trash* | *clown*<br>*hobo*<br>*moron* | *kick*<br>*nuke*<br>*resist* |
| RACISM (NL) | *bakfietsbobo*<br>*kansparel*<br>*zwette* | *bananeplukker*<br>*bruine aap*<br>*makkak* | *beesten*<br>*gespuis*<br>*kakkerlakken* | *geitenneuker*<br>*kamelenpoeper*<br>*kopvod* | *afknallen*<br>*afmaken*<br>*uitroeien* |
| SEXISM | *beta*<br>*chad*<br>*stacy* | *bitch*<br>*faggot*<br>*slut* | *cum dumpster*<br>*femoid*<br>*it* | *nerd*<br>*white knight*<br>*whore* | *beat*<br>*rape*<br>*shoot* |

**Table 4.** Overview of significantly biased HATE words.

## 3.3 Collocation Extraction

Most words are not intrinsically hateful, but can become so in combination with other words (e.g., *bruine* + *aap*), and the variety of hateful word combinations that people can conjure up appears to be infinite. In this regard, collocation extraction (i.e., finding words that co-occur often) is a useful approach. One way to accomplish this is by using part-of-speech tagging, an NLP technique that identifies word types in context, and then counting what adjectives often precede which nouns (for example). In our corpora, this exposes hateful word combinations such as *white devil*, *black bitch*, *bearded carpet kisser* and *fat Muslim smurf*.[24]

A related approach, word embeddings (Mikolov, Yih & Zweig, 2013), is also useful. In brief, each word is mapped to a skip-gram vector that consists of the words that frequently precede or succeed it. This model can then be used to find semantically similar words. For example, we automatically expanded a manual dictionary of 100 hateful German words to 2,000 other hateful words (including word blends, e.g., *refutschies* ≈ "*refusees*") by querying the German Twitter Embeddings (Ruppenhofer, 2018). The resulting dictionary has 75% $F_1$ score and is very easy to interpret, for example by highlighting words for human moderators.

[24] https://docs.google.com/spreadsheets/d/1gVfkxOzLiv47WH506eDseIji96vD2Q4ofFVTiVNUl4c

Figure 6 shows the context of keyword *women* on Incels.me, as a word tree visualization:[25]

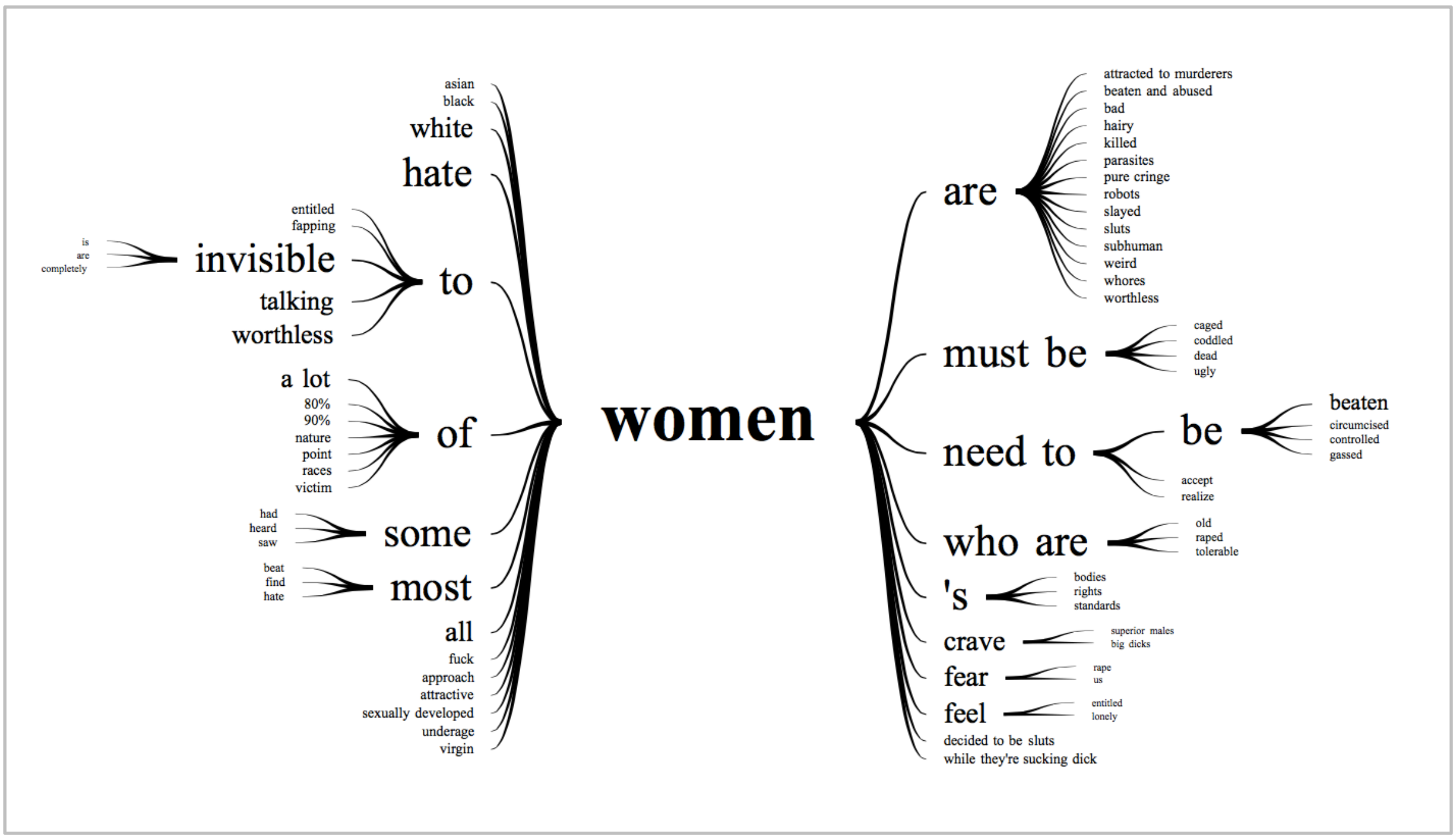

**Figure 6.** Context for keyword *women* in online misogyny.

Skip-gram vectors can also be combined with dimensionality reduction techniques, e.g., to reduce each vector to its two most significant features, after which they can be clustered and represented as a 2-D visualization. We experimented with spherical *k*-means clustering (Hornik et al., 2012) and t-SNE visualization (Maaten & Hinton, 2008). By visually inspecting such clusters, we observed that right-wing extremism mainly focuses on immigration, crime and politics, and that the incel community is ethnically highly heterogeneous, contrary to the popular belief that male supremacy primarily involves white men. Such insights can usually be verified by qualitative analysis, which is more reliable but also more time-consuming.

# 4 Analysis

Not all online hate speech necessarily also involves hateful language. Some instigators simply share news updates (mainly crime reports), but only if these reinforce their worldviews, and then they will do so perpetually. For example, on an average daily basis, German right-wing extremists post 2x more tweets than German politicians did during their election campaign, which can be seen as a form of propaganda. On the other hand, such profiles may also simply show up more in the search queries that we used to collect the corpus.

[25] https://developers.google.com/chart/interactive/docs/gallery/wordtree

In this section, we provide a summary of NLP techniques that we used to profile instigators and their state-of-mind. In general, there is no archetypical instigator, e.g., infighting about the community's belief systems is common. Instead, we argue that **unregulated social media** seems to enable angry persons to band together and stir one another (Sunstein, 2018: 9).

## 4.1 Demographic Profiling

Stylometry refers to NLP techniques that can be used to identify the (anonymous) author of a text, based on the topics discussed or cues in writing style (Argamon, Koppel & Pennebaker, 2009). For example, Pennebaker (2011) has shown that women tend to use more personal pronouns (e.g., *I*, *my*, *we*) to talk about relationships, while men use more determiners (*the*) and quantifiers (*few*, *most*) to talk about objects and concepts. We used the Textgain API[26] to predict the age (75% $F_1$), gender (75% $F_1$), education (80% $F_1$) and personality (60% $F_1$) of the profiles in the collected corpora. While age may vary greatly, there is some consistency in that most users appear to be 20-40 years old. They are more often male, and tend to be somewhat less educated, but this may be an exponent of spelling errors made by non-native speakers, or simply a disregard for proper spelling. Consequently, spelling errors can be a useful predictive cue (e.g., *kakerlakke* ~ *cockroashes*) in combination with other features.

## 4.2 Psychological Profiling

The Linguistic Inquiry and Word Count dictionary (LIWC; Tausczik & Pennebaker, 2010) can be used to identify psychological categories for common words. For example, *cried* is labeled as Affective, Negative, Past, Sadness, and Verb. We applied LIWC to the MISOGYNY corpus and found that incels tend to express more negative emotions (e.g., anger, uncertainty), display more social anxiety, and avoid topics such as family, work, hobbies, goals, and beliefs, which we substantiated with more time-consuming qualitative analysis (see Jaki et al., 2018). The incels feel left behind by society, or as one user states: "Loneliness has followed me my whole life, everywhere. In bars, in cars, sidewalks, stores, everywhere. There's no escape. I'm God's lonely man." The in-group recategorization mechanism of the community is then to blame the "corrupt system" for their lack of self-esteem, social status, job opportunities, and so on. We can wonder about whether this generalizes to instigators in other domains. For example, in Belgium, news coverage about job discrimination based on race or gender surfaces every now and then.[27] In this regard, Wrench (2016) offers more in-depth perspectives.

[26] https://textgain.com
[27] http://www.brusselstimes.com/opinion/11647/the-integration-process-in-belgium-bureaucracy-inefficiencies-and-incompetence

## 4.3 Sentiment Analysis

Many users that post hate speech are angry, for one reason or another. This is manifested in a negative intensity in their language use, e.g., adversaries are not simply *parasites*, but *fucking parasites*, and what they say or do is not just *nonsensical*, but *UTTER NONSENSE*, and *enraging*, *infuriating*, *unforgivable*, etc., reinforced with exclamation marks, all caps, and emojis such as the angry face, the vomit face, or the fist punch (see Figure 7).

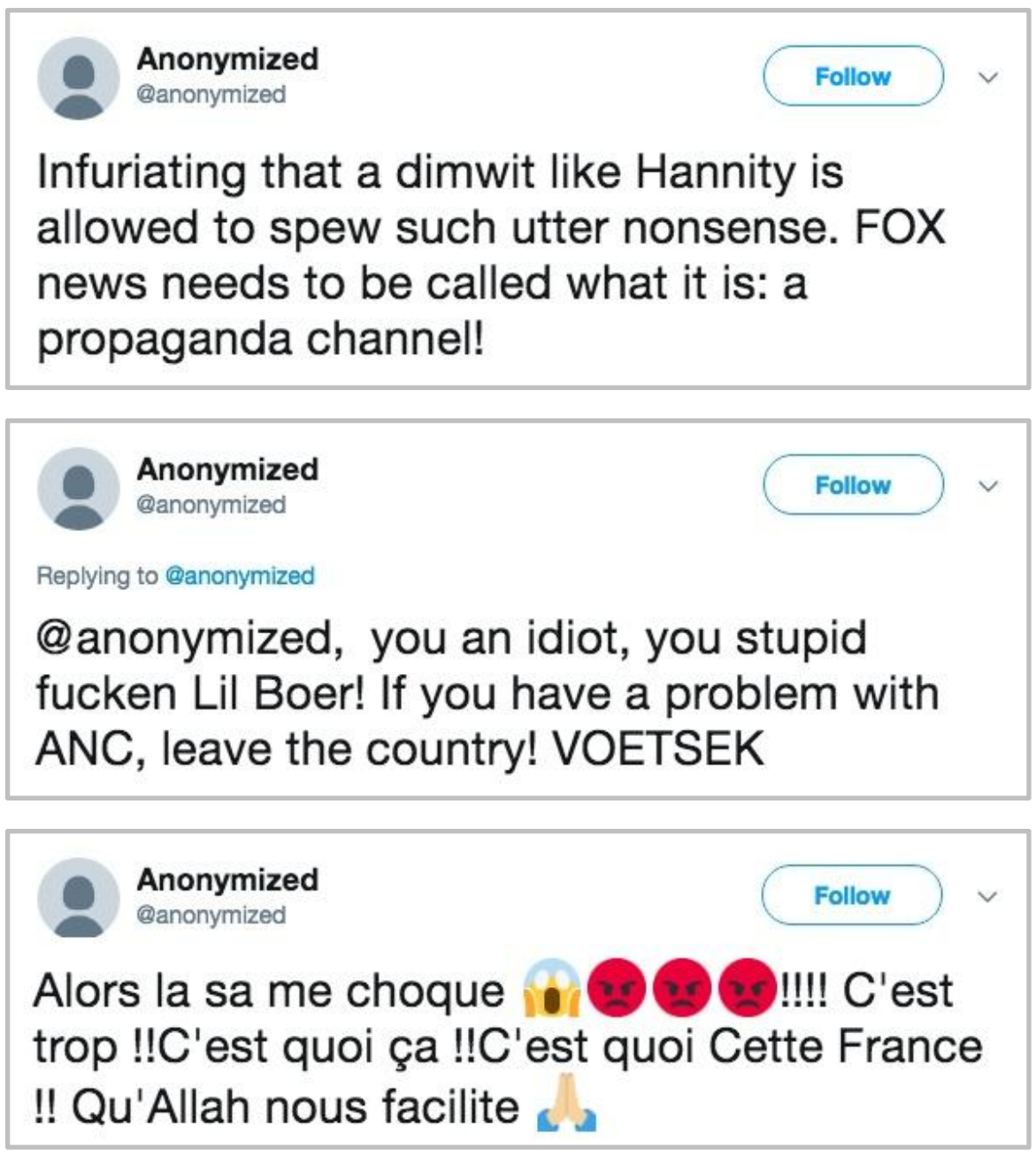

**Figure 7.** Examples of negativity markers.

Sentiment analysis refers to NLP approaches that detect whether a text is objective (facts) or subjective (opinions), and if it is subjecttive, if it is positive or negative (Liu, 2012). This is often accomplished with a dictionary of adjective scores (*good* +0.5, *bad* –0.5) or models trained on reviews and star ratings. Since negative adjectives occur frequently in hate speech, the sentiment score predicted by these systems is usually negative as well, which can be a good predictive cue. Figure 8 shows the average polarity in random Wikipedia articles and tweets vs hate speech.

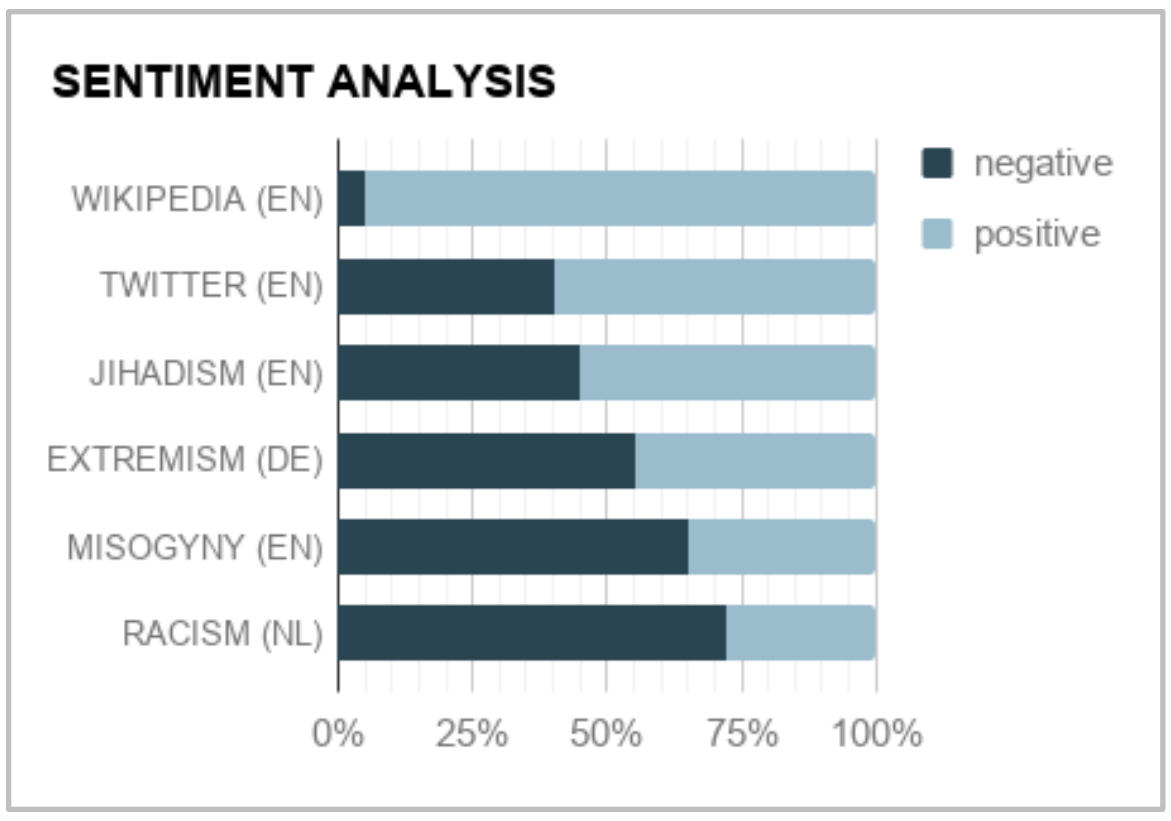

**Figure 8.** Sentiment polarity comparison.

Using open source sentiment dictionaries that report about 80% $F_1$ score (e.g., De Smedt & Daelemans, 2012b), we found that jihadist hate speech is more negative than random tweets, that German right-wing extremism tweets are even more negative, as well as the majority of incel misogyny (~65%) and Dutch racism (~75%). In the South African domain, sentiment analysis was less accurate, for example because many open source dictionaries assign a negative score to *apartheid*, which occurs frequently in neutral South African political debate (e.g., "We explain what is Orania, how it was established after the end of *apartheid*" is mislabeled as negative). We did note that negative sentiment is less common in tweets that take a defensive view (i.e., pro Orania) than in tweets with and offensive view (against Orania).

## 4.4 Network Analysis

In cases where we need to find the "influencers", i.e., the users that post original content and that are frequently cited, we found network analysis to be useful. Essentially, messages in the corpus are mapped to an X CITES Y graph, for example by extracting the authors' usernames (X) and `@retweeted` usernames in their messages (Y). Mathematical techniques from graph theory can then be applied to gain insight. In the JIHADISM corpus, we were able to discern between fans and insurgents by using eigenvector centrality (cf. Page et al., 1999), which assigns a greater weight to nodes that are indirectly linked more often. This exposed a number of usernames belonging to ISIS recruiters that have later been convicted in court. Figure 9 shows an example (some usernames belong to neutral experts and news media).

The main problem is that usernames usually do not reveal the author's real name, so when the profile is suspended, a new profile appears, and with current NLP techniques it is hard to detect whether such profiles are one and the same user. For more specialized methods that have been used to examine terrorist networks, see Benigni, Joseph & Carley (2017).

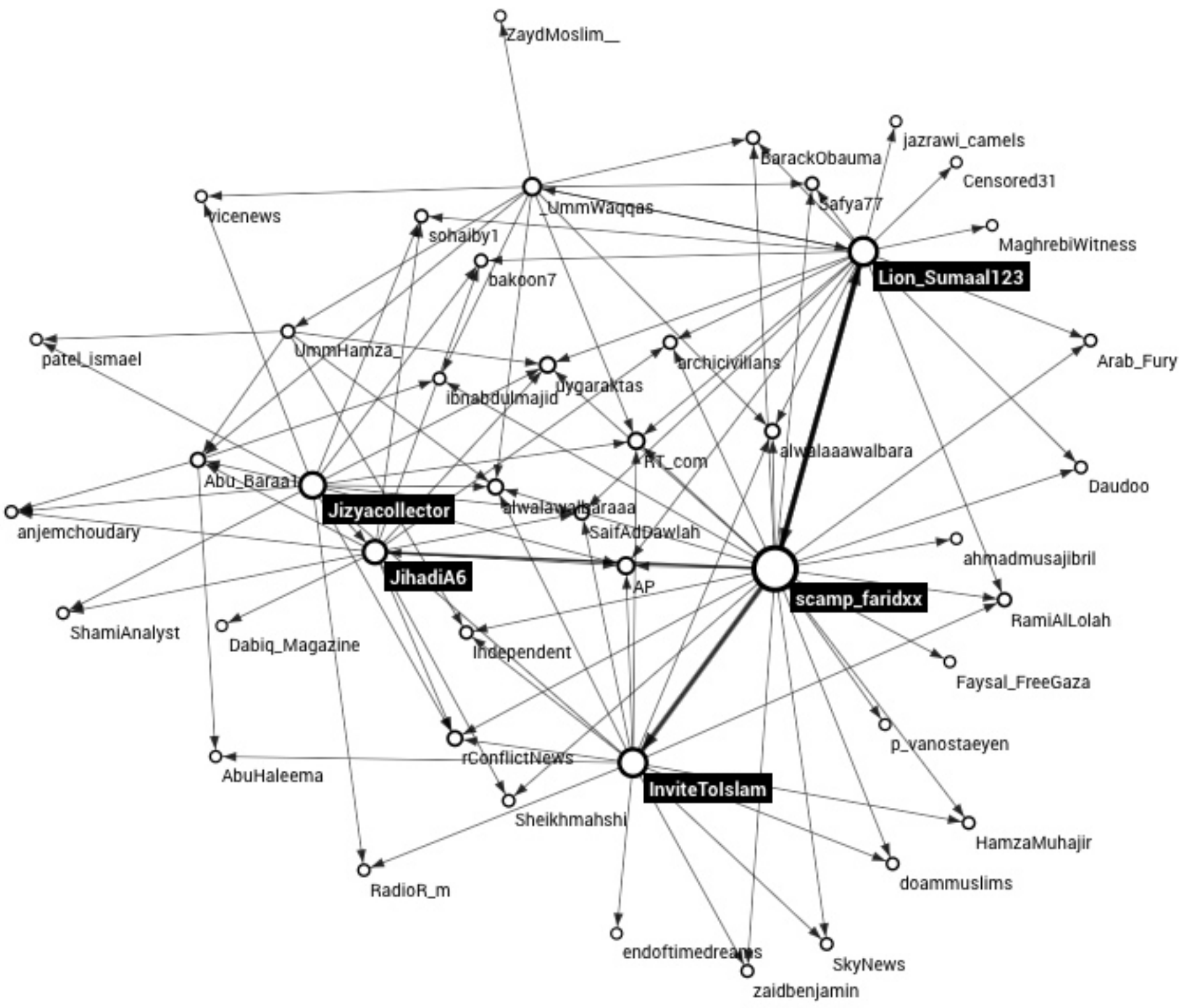

**Figure 9.** Example hate cluster, pruned by eigenvector centrality.

# 5 Discussion

In this report, we have outlined general findings on online hate speech in different languages and domains. We demonstrated how detecting such content is theoretically viable, with an in-domain accuracy of about 80-95%. We believe that our study of the varied, yet often similar corpora can be relevant to future research on online hate speech, and may also have practical benefits, such as providing training material for detection and de-escalation of conflicts online. However, automatic detection of hate speech is challenging for a number of legal and ethical reasons. For one, hate speech laws often tend to be vague, in an attempt to capture a wide range of problematic content without infringing on the freedom of expression of private citizens. Without a well-defined legal framework, what exactly are we detecting? What is the threshold from sharing a personal opinion in a fit of anger to defamation and incitement? Also, the new GDPR[28] regulations now curtail the kind of data that can be collected in the EU ("right to be forgotten").[29] Countering online hate speech using automated techniques may well be outside the legal realm, despite all good intentions.

Hate speech can target women and men alike, and any ethnicity, ideology or religion, so it is hard to come up with a linguistic definition. There is no standardized "list of bad words", and if there is, then perpetrators are very creative in coining new offensive terminology. Online machine learning algorithms may be useful to continuously learn to detect evolving rhetoric, but even if such approaches are successful, perpetrators can always argue that they were only joking. Finally, developing useful technology tools does not mean that IT companies will adopt it, e.g., Facebook will not remove Holocaust denial.[30]

[28] http://en.wikipedia.org/wiki/GDPR
[29] http://en.wikipedia.org/wiki/Right_to_be_forgotten
[30] http://www.reuters.tv/v/hzL/2018/07/18/zuckerberg-gets-heat-for-holocaust-comment